# Novel Approach to Use HU Moments with Image Processing Techniques for Real Time Sign Language Communication

**Matheesha Fernando**                                    *matheemit@gmail.com*
*Department of Industrial Management,*
*University of Kelaniya*
*Sri Lanka*

**Janaka Wijayanayake**                                    *janaka@kln.ac.lk*
*Department of Industrial Management,*
*University of Kelaniya*
*Sri Lanka*

## Abstract

Sign language is the fundamental communication method among people who suffer from speech and hearing defects. The rest of the world doesn't have a clear idea of sign language. "Sign Language Communicator" (SLC) is designed to solve the language barrier between the sign language users and the rest of the world. The main objective of this research is to provide a low cost affordable method of sign language interpretation. This system will also be very useful to the sign language learners as they can practice the sign language. During the research available human computer interaction techniques in posture recognition was tested and evaluated. A series of image processing techniques with Hu-moment classification was identified as the best approach. To improve the accuracy of the system, a new approach; height to width ratio filtration was implemented along with Hu-moments. System is able to recognize selected Sign Language signs with the accuracy of 84% without a controlled background with small light adjustments.

## 1. INTRODUCTION

More than 40 million of the world population suffers from speech impairments. Sign Language is the mode of communication of speech and auditory impaired people. In today's world communication facilities provided for such people are very less. Generally people in the society don't know about sign language. An ordinary person to communicate with speech and auditory impaired person, a translator is usually needed to convert the sign language into natural language and vice versa. This is an inflexible way of communication and restricts communication between speech and hearing impaired people and ordinary people.

The objective of the research is to develop a technology that can act as an intermediate flexible medium for speech and hearing impaired people to communicate amongst themselves and with other individuals to enhance their level of learning / education.

The research outcome will facilitate speech and hearing impaired people to communicate with the world in a more comfortable way where they can easily get what they need from the society and also can contribute to the well-being of the society. Furthermore, another expectation is to use the research outcome as a learning tool of sign language where learners can practice signs.

The rest of this paper is organized as follows. Section 2 briefly introduces the area of study with related work in sign language recognition and human computer interaction. In section 3 the proposed methodology for feature extraction is described in detail. Section 4 describes the proposed methodology for classification with the experiments results. Section 5 reveals the limitations and drawbacks of the proposed approach. In section 6 possible future improvements are introduced and in section 7 the conclusion of the research findings is presented.

## 2. RELATED WORK

With the development of information technology new areas of human computer-interaction are emerging. There, human gesture plays a major role in the field of human computer interaction. Hand gestures provide a separate complementary modality to speech for expressing ones ideas. As sign language is a collection of gestures and postures any effort in sign language recognition falls in the field of human computer interaction and image processing.

There are two types of approaches commonly used to interpret gestures for human computer interaction. First category, Data Glove based approach relies on electromechanical devices attached to a glove for digitizing hand and finger motions into multi-parametric data. The major problems with that approach are; it requires wearing the devices which are expensive and also cause less natural behaviors [1]. The second category, Vision based approach is further divided into two categories as3D hand model based approach and appearance based approach. 3D hand model based approach relies on the 3D kinematic hand model and tries to estimate the hand parameters by comparison between the input images and the possible 2D appearance projected by the 3D hand model. The appearance based approach uses image features to model the visual appearance of the hand and compare these parameters with the extracted image features from the video input [1, 2].

There have been a number of research efforts on appearance based methods in recent years. Freeman and Roth [3] presented a method to recognize hand gestures based on pattern recognition technique employing histograms of local orientation. There are some different types of gestures with similar orientation histograms which cannot be separately recognized by this method. This is one of the major drawbacks of Freeman and Roth [3] proposed method. Orientation histogram method is most appropriate for close-ups of the hand and also it doesn't provide a method to recognize hand itself.

Roth and Tanaka [4] have presented a way of classifying images based on their moments with more concern on recognizing shapes of the static postures. Serge and Malik [5] have discussed an approach to measure similarity between shapes and exploit it for object recognition where they attach a descriptor – shape context to each point. The corresponding points in similar shapes will have similar shape contexts. By that they compute the sum of matching errors between corresponding points. This requires more computational power. Flusser [6] presented a survey on Moment Invariants in Image Analysis on object recognition /classification methods based on image moments. The paper presents a general theory to construct invariant classification methods. Nianjun Liu and Brian C. Lovell [7] have presented an approach to Hand Gesture Extraction by Active Shape Models where a set of feature vectors will be normalized and aligned and then trained by Principle Component Analysis (PCA). Chen Chang [8] came up with an approach for Static Gesture Recognition by recognizing static gestures based on Zernike moments (ZMs) and pseudo-Zernike moments (PZMs). This approach takes a step toward extracting reliable features for static gesture recognition. Zhang et al. [9] suggested a Fast Convex Hull Algorithm for Binary Image for pattern recognition. The recognition is achieved by computing the extreme points, dividing the binary image into several regions, scanning the region's existing vertices dynamically, calculating the monotone segments, and merging these calculated segments. Deng and Jason, [10] came up with an idea of shape context based matching with cost matrix for real time Hand Gesture Recognition. There, they translate the edge elements of image shape to a group of feature points with N value.

Erdem Yörük et al. [11] have suggested a method for person recognition and verification based on their hands. There at preprocessing stage of the algorithm, the silhouettes of hand images are registered to a fixed pose, which involves both rotation and translation of the hand and the individual fingers separately. Two feature sets, Hausdorff distance of the hand contours and independent component features of the hand silhouette images have been comparatively assessed. The system is not implemented in real time as it is tested based on the images of the hands captured by a scanner.

In appearance based method, feature extraction and classification are the major components that take part in. Under feature extraction Hand contour, Multi scale color features, Scale Invariant Feature Transform (SIFT), Speeded Up Robust Feature(SURF), Haar-like features, Histograms of Oriented Gradients (HOG), Local Orientation Histogram, Hough Transform (HT) are some of available research findings in efficient feature extraction. Qing Chen [12] has used a combined approach for recognition using two levels. Haar-like features were used to hand posture detection and tracking. AdaBoost learning algorithm was used to construct an accurate cascade of classifiers by combining a sequence of weak classifiers and to recognize different hand postures with parallel cascades. Yikai Fang et al. [13] have suggested a combined approach of co-training haar and HOG features simultaneously reduces the required training samples.

## 3. METHODOLOGY

As shown in Figure 1, this research is focused on a system that can convert a video signal (processed as sequence of images) into a sequence of written words (text) and voice in real time. Further, in the other way round it can provide a sign demonstration for a given text.

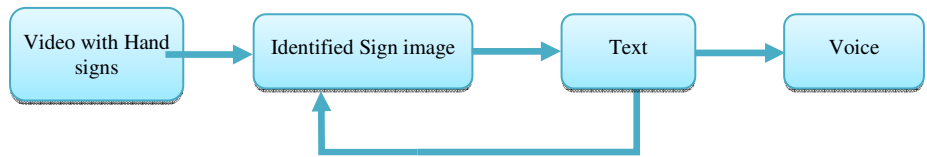

**FIGURE 1:** The Idea.

Robustness, Computational Efficiency and User's Tolerance were the important challenges need to be considering in conducting this research project. Overall methodology followed for the sign language classification is summarized in figure 2.

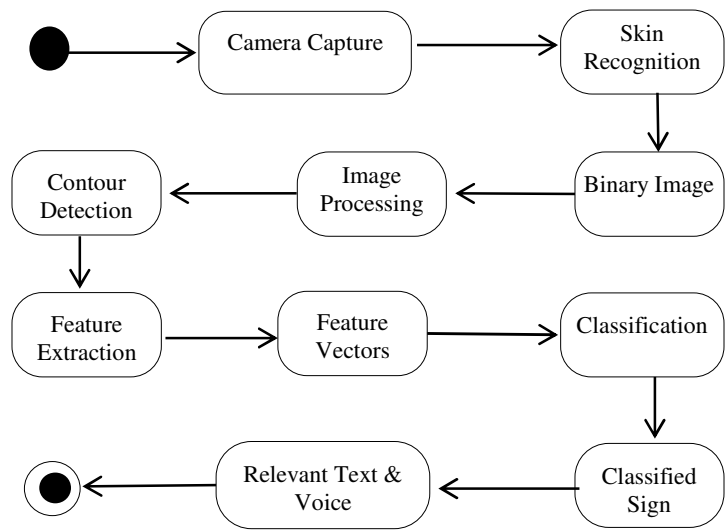

**FIGURE 2:** Classification Process.

In this approach the movement of the hand is recorded by a camera and the input video is decomposed into a set of features by taking individual frames into account. The video frames contain background pixels other than the hand, as a hand will never fill a perfect square. These pixels have to be removed as they contain arbitrary values that introduce noise into the process.

System captures images containing the hand signs from a video first and then identifies the skin colored area and gets the hand separated out from the rest of the background. Thereafter, a binary mask for skin pixels is constructed. Then the hands are isolated from other body parts as well as background objects. The basic idea is to get the real time image and then extract the predefined features and compare the feature vectors against the features extracted from stored template sign images. Sign database has the text and voice relevant to the meaning of each sign. When the real time sign is matched with a stored sign, the system will provide the relevant text and voice.

### 3.1. Image Capture Phase
Only a web camera was used to make this system a low cost, affordable application and that made the capturing process simple. As the normal web camera images are very poor in quality it requires more image processing techniques to deal with the issue.

### 3.2. Hand Segmentation
The hand must be localized in the image and segmented from the background before recognition. Color is the selected feature due to various reasons such as its computational simplicity, its invariant properties regarding the hand shape configurations and due to the human skin-color characteristic values. The method is based on a more rule based model of the skin-color pixels distribution. Since people have different hand colors it is incorrect to rely on the intensity. Therefore, it is required to convert the RGB representation to an intensity free color space model. Hue-Saturation-Value (HSV) representation and YCrCb color space are the best options available [14]. In YCrCb, Y is the luminance component and CB and CR are the blue-difference and red-difference chroma components. YCC is simply created from RGB [15].

The advantage of YCC over RGB is that it can reduce resolution of the color channels without altering the apparent resolution of the image, since the main Y (luminance) signal is untouched. YCrCb derived from the corresponding RGB space is as follows,
$Y = 0.299R + 0.587G + 0.114B$
$Cb = 0.564(B - Y) + 1/2$ full scale
$Cr = 0.713(R - Y) + 1/2$ full scale

After a number of iterations of experiment, the suitable color ranges for human skin in both HSV and YCrCb were figured out as follows,
hsv_min = (0, 30, 60)
hsv_max = (30, 150, 255)
YCrCb_min = (0, 131, 80)
YCrCb_max = (255, 185, 135)

In a color image there is more information to be processed. The processing part is more complicated and time consuming. Therefore, binarizing (converting the image to black and white) is more desirable, since it reduces the amount of information contained in the source image. Thresholding can be used to binarize the image. Figure 3 shows the effect of binarizing.

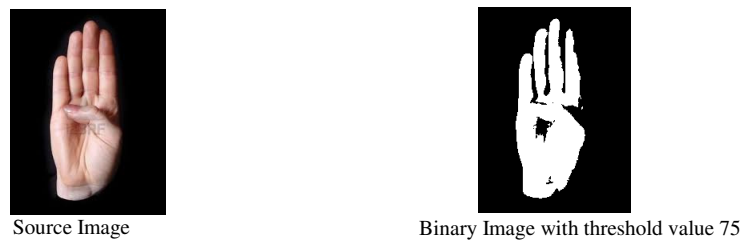

Source Image                    Binary Image with threshold value 75

**FIGURE 3:** Binary Image.

### 3.3. Image Pre-processing Phase

Image processing techniques were used to improve the quality of the image. Morphological transformations were used to get a more clear hand image. The basic morphological transformations are called dilation and erosion, and they arise in a wide variety of contexts such as removing noise, isolating individual elements, and joining disparate elements in an image. The skin recognized image was processed with both dilation and erosion to get the image clearer as shown in figure 4.

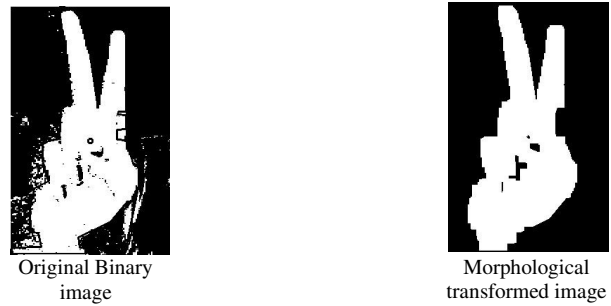

Original Binary image

Morphological transformed image

**FIGURE 4:** Morphological Transformation.

### 3.4. Contour Detection

Detecting contours of a hand image means finding edges that have high contrast pixels than its neighbors. In the real world a hand image may have dark places and shadows not only at the ends but also at the middle of the hand. It will result in a set of contours in various lengths. Filtering the outer contour is done under feature extraction phase.

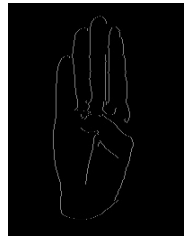

**FIGURE 5:** Contour Detected.

### 3.5. Feature Extraction

In the feature extraction phase what is most important is to get possible precise features as output. Features selected for classification are hand contour, orientation histogram, convex hull, convexity defects and hu moments.

The convex hull for the hand shape was computed and the poly line minimum area covering box and the rectangle was identified. Then the center of the box was computed. The Region of Interest (ROI) of the image was set to the minimum rectangle. In the algorithm Region of Interest (ROI) plays a major role as it is the area subjected to matching.

## 4. CLASSIFICATION

At the first stage of research, practicing and testing on image processing techniques were done and through that the investigations were carried out to identify the best methods and algorithms for sign language recognition. These attempts were focused to find out the best matching algorithm for sign language recognition. Different researched methods and algorithms in the literature of human computer interaction such as template matching and shape context matching were implemented and tested with modifications to apply them into sign language recognition domain. Normalized template matching was tested with equation,

$$(x, y) = \sqrt{\sum_{x',y'} T(x',y') . \sum_{x',y'} I(x+x', y+x')^2} \qquad (1)$$

Contour matching was tested where it compared the number of points match at the sequences. To make the contour features more powerful, filtration of hand out line contour was used. The biggest contour by the area that each contour covers was computed for filtering process. As the experiment results shows the biggest contour is very much close to hand out line.

Shape match with contour was introduced considering the outer contour as a shape and did the shape matching. Then defects computation for the given real time image was done, where it calculates the total area which is differed from the template, and that was used for classification.

Histogram based classification was implemented with binary image, as histograms can be used to represent the color distribution of an object, an edge gradient template of an object etc. [3]. Here the histogram was generated for the given binary images and the histograms were normalized before matching. As the whole image is used for histogram construction the computational efficiency is very less and the real time concept didn't actually work with this model. As a solution, histograms were computed for contours considering the drawbacks of the previous method, with the idea to reduce the image data for histogram construction. Histogram for a contour is shown in figure 6.

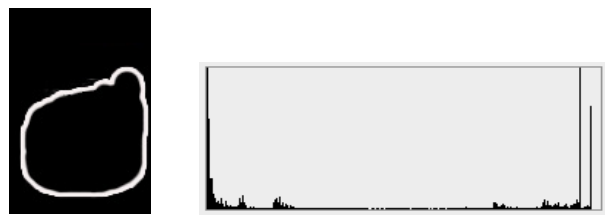

**FIGURE 6:** Histograms and Contour.

Then the moment based recognition was implemented and tested. Moment based recognition showed more accuracy in identifying the signs than all other attempts mentioned before. In general, (p, q) moment of a contour can be defined as,

$$m_{p,q} = i=1 \sum n \; I(x,y) x^p y^q \qquad (2)$$

Here p is the x-order and q is the y-order, whereby order means the power to which the corresponding component is taken in the sum just displayed. The summation is over all of the pixels of the contour boundary (denoted by n in the equation) [16].

### 4.1. Hu Moments
Hu moments use the central moments where it is basically the same as the moments just described except that the values of x and y used in the formulas are displaced by the mean values.

$$\mu_{p,q} = \sum_{i=0}^{n} I(x,y) \left(x - x_{avg}\right)^p \left(y - y_{avg}\right)^q \qquad (3)$$

Where x(avg) = m10/m00 and y(avg)=m01/m00

The idea here is that, by combining the different normalized central moments, it is possible to create invariant functions representing different aspects of the image in a way that is invariant to scale and rotation. Seven hu moment calculations by Hu (1962) are shown below [16].

$$I_1 = \eta_{20} + \eta_{02}$$
$$I_2 = (\eta_{20} - \eta_{02})^2 + 4\eta_{11}^2$$
$$I_3 = (\eta_{30} - 3\eta_{12})^2 + (3\eta_{21} - \eta_{03})^2$$
$$I_4 = (\eta_{30} + \eta_{12})^2 + (\eta_{21} + \eta_{03})^2$$
$$I_5 = (\eta_{30} - 3\eta_{12})(\eta_{30} + \eta_{12})[(\eta_{30} + \eta_{12})^2 - 3(\eta_{21} + \eta_{03})^2] + (3\eta_{21} - \eta_{03})(\eta_{21} + \eta_{03})[3(\eta_{30} + \eta_{12})^2 - (\eta_{21} + \eta_{03})^2]$$
$$I_6 = (\eta_{20} - \eta_{02})[(\eta_{30} - \eta_{12})^2 - (\eta_{21} + \eta_{03})^2] + 4\eta_{11}(\eta_{30} + \eta_{12})(\eta_{21} + \eta_{03})$$
$$I_7 = (3\eta_{21} - \eta_{03})(\eta_{30} + \eta_{12})[(\eta_{30} + \eta_{12})^2 - 3(\eta_{21} + \eta_{03})^2] - (\eta_{30} - 3\eta_{12})(\eta_{21} + \eta_{03})[3(\eta_{30} + \eta_{12})^2 - (\eta_{21} + \eta_{03})^2].$$

As hu moments are rotation invariant the conflicts is aroused between the shapes that have similar shapes in different angles as shown in the figure 7.

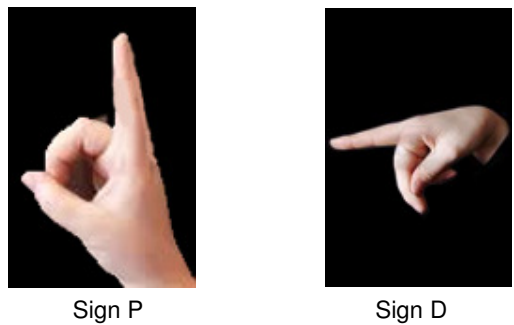

Sign P                    Sign D

**FIGURE 7:** Problem of Rotation Invariant.

If sign D is 90 degrees rotated anti clockwise or sign P is 90 degrees rotated clockwise then these two signs have a very similar shape. Therefore, for a rotation invariant algorithm such as hu moments or histograms, these shapes are the same. In [17] the hidden problem behind the unsuccessfulness of Hu moment can be reason out with above discovery. To overcome this problem it requires a new algorithm or any other method.

The solution suggested by researchers is hu moments with height to width ratio filtration. Considering the drawbacks of using Hu-moments, this proposed method works fairly well. As the research found out, problems occur between the signs that are same when rotated. The method proposed by this research gives better performance in recognizing uniqueness between the signs that are rotationally equal.

### 4.2. Height to Width Ratio
First for a given sign the Region of Interest (ROI) is identified using image processing and skin recognition methods. Identification of ROI is shown in figure 8. Thereafter, for the identified hand area the height to width ratio is calculated as shown in table 1.

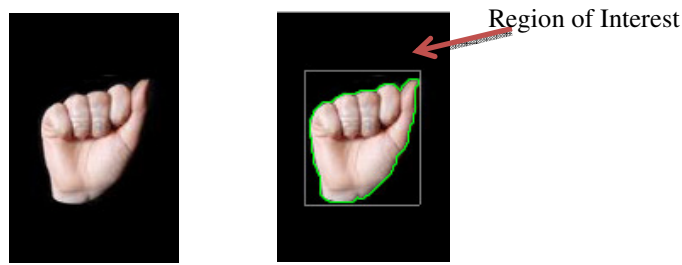

Region of Interest

**FIGURE 8:** Region of Interest Identification.

| Sign | Image | Height/Width | Ratio |
|------|-------|-------------|-------|
| A | 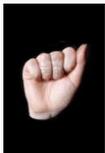 | 73 / 68 | 1.07352941176471 |
| B | 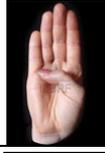 | 230 / 98 | 2.3469387755102 |
| C | 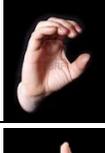 | 124 / 94 | 1.31914893617021 |
| D | 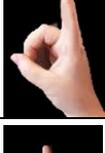 | 152 / 108 | 1.40740740740741 |
| L | 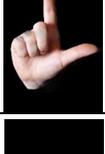 | 247 / 245 | 1.00816326530612 |
| P | 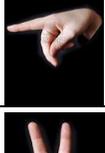 | 169 / 235 | 0.719148936170213 |
| V | 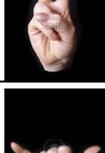 | 141 / 63 | 2.23809523809524 |
| Y | 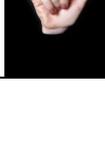 | 55 / 57 | 0.964912280701754 |

**TABLE 1:** Height to Width Ratio.

When the ratio is considered, total algorithm remains to be scale invariant and also the rotation is invariant up to certain point. That means as the consideration is on whether the ratio => 1 or ratio <$10^0$ to $30^0$ deviation of the sign won't affect much on the height to width ratio. However, when the rotation is more than $30^0$, the new method can recognize the sign correctly. Hu moments classification with height to width ratio is found out to be the best matching method for sign language recognition.

## 5. RESULTS AND DISCUSSION

In finding the actual output of hu moment matching, table 2 display some real time generated hu moment matching result values.

| Sign | Template image | Real time image | Result values |
|------|---------------|-----------------|---------------|
| A | 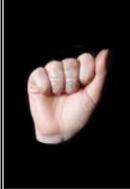 | 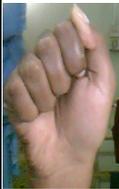 | A - 0.08956746514951<br>V - 0.27577140714475<br>L - 0.34207054932742<br>B - 0.24181056482163<br>D - 0.37176152384964<br>C - 0.18130743044722<br>P - 0.25942242245184<br>Y - 0.43210990521291 |
| B | 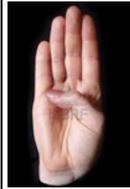 | 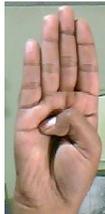 | A - 0.54202431872088<br>V - 0.20274582657002<br>L - 0.77298229967616<br>B - 0.08136118898609<br>D - 0.48160990559653<br>C - 0.32445697847339<br>P - 0.38891345418583<br>Y - 0.53140628754324 |
| C | 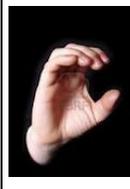 | 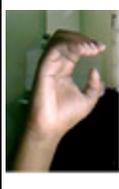 | A - 0.22470657026369<br>V - 0.26110625710516<br>L - 0.40813802027182<br>B - 0.27948067338527<br>D - 0.27798597325585<br>C - 0.05910328442404<br>P - 0.14887830150346<br>Y - 0.34734240316219 |
| D | 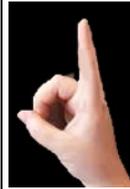 | 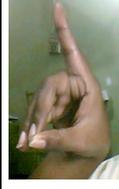 | A - 0.49651217137801<br>V - 0.34664529950186<br>L - 0.72065353243200<br>B - 0.55167658949923<br>D - 0.24146058292755<br>C - 0.32495304940430<br>P - 0.25311353806331<br>Y - 0.42539378543313 |
| V | 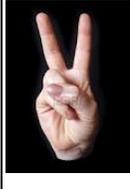 | 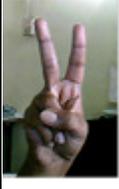 | A - 0.77538997758076<br>V - 0.15136323806656<br>L - 1.0413004951077<br>B - 0.32666217539152<br>D - 0.47280905471127<br>C - 0.52194655877205<br>P - 0.48663355955186<br>Y - 0.69101854910652 |

**TABLE 2:** Hu Moment Calculation Results.

Hu moments alone was invariant to scale and rotation and succeeded in identifying signs that are not identical in shape than other classification algorithms. But when the sign collection increased with different signs that has same shape with different orientation, Hu moment based classification tend to mislead.

Testing was conducted with 50 signs from 10 users where 5 signs (A, B, C, D and V) from each individual. 8 signs were stored as templates including testing 5 signs and 3 other signs (A, B, C, D, L, P, V and Y). When the test was conducted without the height to width ratio addition to the hu algorithm the accuracy was 76% as Sign D was conflicted with Sign P. Exact similar behavior has been reported in the research by Rodriguez et al. [17]. When the testing was conducted with suggested approach, out of 50 signs only 8 signs failed to identify clearly. That gives an accuracy level of 84%. This research clearly identifies that hu moments can be used to classify sign language recognition by incorporating controlled rotation variance.

## 6. LIMITATIONS AND DRAWBACKS

As this is a real time application, process is highly depends on the user's behavior. Being an appearance based solution, the system is very sensitive to light condition of the environment. In this research only the postures of sign language signs were considered based on the idea that postures are fair enough to identify a certain level of sign language. The spatial information for the hand signs was not considered for classification. There can be some occasions where the same sign posture is presented for different meanings with spatial differences.

## 7. FUTURE IMPROVEMENT AND EXTENSIONS POSSIBLE

Sign language is a very vast area with its grammar and literature component. For the research mainly ASL (American Sign Language) alphabet is used as the test cases. In future, this project can be advanced to convert words, phrases and simple sentences. Further, this project only looks at the hand postures as it is the major part of sign language. In future research, this can be extended to identify hand gestures with spatial information. There a state machine for gesture recognition can be implemented with a machine learning algorithm where it will result in an intelligent sign language recognition system. Movements of the hand sign can be detected by taking the absolute difference of the center of ROI box. As a further improvement, facial expressions can be recognized to get more sophisticated combined idea.

## 8. CONCLUSION

The proposed research has dealt with the sign language gesture/posture and identified the sign and converted that sign into text and voice. The design phase and development phase were conducted together using a more agile software architecture where it improves with more added functionalities and algorithm developments with the evolution of system design. Main part of the design was to find a best fitting algorithm for sign language posture recognition. A series of image processing techniques with Hu-moment classification was identified as the best approach. To improve the accuracy of the system, a new approach; height to width ratio filtration was implemented along with Hu-moments. System is able to recognize selected Sign Language signs with the accuracy of 84% without a controlled background with small light adjustments.

## 9. REFERENCES

[1]   C. Harshith, R.S. Karthik, M. Ravindran, M.V.V.N.S Srikanth, N. Lakshmikhanth, "Survey on various gesture recognition Techniques for interfacing machines based on ambient intelligence", *International Journal of Computer Science & Engineering Survey (IJCSES)* Vol.1, No.2, November, 2010

[2]   Q. Chen, N.D. Georganas, E.M. Petriu, "Real-time Vision-based Hand Gesture Recognition Using Haar-like Features", *Instrumentation and Measurement Technology Conference*, 2007

[3]   W. T. Freeman and M. Roth, "Orientation histograms for hand gesture recognition", *IEEE Intl. Wkshp. on Automatic Face and Gesture Recognition*, Zurich, June, 1995

[4]   M. Roth, K. Tanaka, C. ssman, W. Yerazunis, "Computer Vision for Interactive Computer Graphics", *IEEE Computer Graphics and Applications*, May-June, 1998, pp. 42-53

[5] S. Belongie, J. Malik, "Shape Matching and Object Recognition Using Shape Contexts", *IEEE Transactions on pattern analysis and machine intelligence*, Vol.24, No.24, 2002

[6] J. Flusser, "Moment Invariants in Image Analysis", *World Academy of Science, Engineering and Technology*, 2005

[7] N. Liu, B. C. Lovell, "Hand Gesture Extraction by Active Shape Models", *Digital Image Computing: Techniques and Applications, DICTA '05. Proceedings*, 2005

[8] C. chang, J. chen, W. Tai and C. Han, "New Approach for Static Gesture Recognition", *Journal of information science and engineering*, 2006

[9] X. Zhang and Z. Tang, J. Yu, M. Guo, "A Fast Convex Hull Algorithm for Binary Image", *Informatica Oct2010*, Vol. 34 Issue 3, pp.369, 2010

[10] L. Y. Deng, J. C. Hung, H. Keh, K. Lin, Y. Liu, and N. Huang, "Real-time Hand Gesture Recognition by Shape Context Based Matching and Cost Matrix", *Journal of Networks*, Vol. 6, No. 5, May 2011

[11] E. Yörük, E. Konukoğlu, B. Sankur, "Shape-Based Hand Recognition", IEEE transactions on image processing, vol. 15, no. 7, July 2006, 2006

[12] Q. Chen, "Real-Time Vision-Based Hand Tracking and Gesture Recognition", Doctoral Dissertation, University of Ottawa, 2008

[13] Y. Fang, J. Cheng, J. Wang, K. Wang, J. Liu, H. Lu , "Hand Posture Recognition with Co-Training", *19th International Conference on Pattern Recognition*, 2008

[14] G. Kukharev, A. Nowosielski, "Visitor Identification - Elaborating Real Time Face Recognition System", *WSCG Short Communication papers proceedings,* 2004, pp. 157-164

[15] X. Zabulisy, H. Baltzakisy, A. Argyroszy, "Vision-based Hand Gesture Recognition for Human-Computer Interaction", *World Academy of Science, Engineering and Technology*, 2006

[16] M.K. Hu, "Visual pattern recognition by moment invariants", Information Theory, *IRE Transaction*s, 1962,  pp. 179-187

[17] K.C.O. Rodriguez, G.C. Chavez, D. Menotti, "Hu and Zernike Moments for Sign Language Recognition", *Computing Department, Federal University of Ouro Preto*, Brazil, 2012